\documentclass[10pt,twocolumn,letterpaper]{article}

\usepackage[pagenumbers]{cvpr} 

\usepackage[accsupp]{axessibility} 

\definecolor{cvprblue}{rgb}{0.21,0.49,0.74}
\usepackage[pagebackref,breaklinks,colorlinks,allcolors=cvprblue]{hyperref}

\def\paperID{2} 
\def\confName{CVPR}
\def\confYear{2025}

\title{Few-Shot Adaptation of Grounding DINO for Agricultural Domain}

\author{Rajhans Singh, Rafael Bidese Puhl, Kshitiz Dhakal, Sudhir Sornapudi \\
Corteva Agriscience, Indianapolis, USA\\
{\tt\small \{rajhans.singh, rafael.bidesepuhl, kshitiz.dhakal-1, sudhir.sornapudi\}@corteva.com}
}

\begin{document}

\twocolumn[{%
\renewcommand\twocolumn[1][]{#1}%
\maketitle

\begin{center}
\parbox{\textwidth}{
    \captionsetup{type=figure}
    \vspace{-0.3in}
    \includegraphics[width=\textwidth]{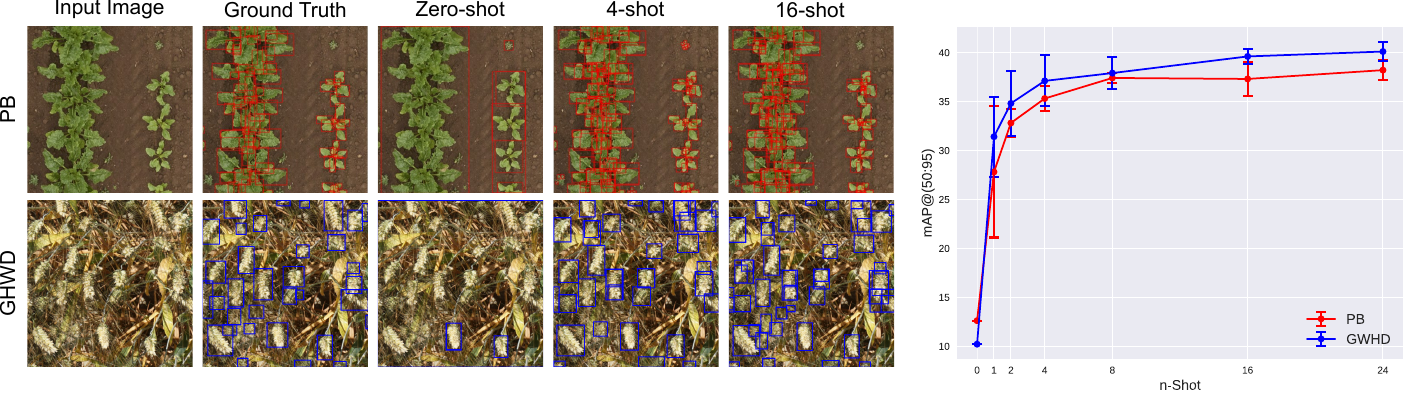}
    \vspace{-0.25in}
    \captionof{figure}{Left: zero-shot vs our few-shot (4-shot, 16-shot) using Grounding DINO on Wheat Head (GWHD) \cite{david2023global} and PhenoBench (PB) \cite{weyler2024phenobench}. Zero-shot fails in cluttered/occluded environments, whereas our few-shot outperforms significantly. Right: mAP of our few-shot approach increases with more training images.}
    \label{fig:teaser}
    \vspace{-0.05in}
    }
\end{center}%
}]
    
    \maketitle
 
    \thispagestyle{empty}
    
    \begin{abstract}
Deep learning models are transforming agricultural applications by enabling automated phenotyping, monitoring, and yield estimation. However, their effectiveness heavily depends on large amounts of annotated training data, which can be labor and time intensive. Recent advances in open-set object detection, particularly with models like Grounding-DINO, offer a potential solution to detect regions of interests based on text prompt input. Initial zero-shot experiments revealed challenges in crafting effective text prompts, especially for complex objects like individual leaves and visually similar classes. To address these limitations, we propose an efficient few-shot adaptation method that simplifies the Grounding-DINO architecture by removing the text encoder module (BERT) and introducing a randomly initialized trainable text embedding. This method achieves superior performance across multiple agricultural datasets, including plant-weed detection, plant counting, insect identification, fruit counting, and remote sensing tasks. Specifically, it demonstrates up to a $\sim24\%$ higher mAP than fully fine-tuned YOLO models on agricultural datasets and outperforms previous state-of-the-art methods by $\sim10\%$ in remote sensing, under few-shot learning conditions. Our method offers a promising solution for automating annotation and accelerating the development of specialized agricultural AI solutions.
\end{abstract}    
    \section{Introduction}
\label{sec:introduction}


Deep learning based object detection \cite{ren2015faster, carion2020end, redmon2016you} and segmentation \cite{he2017mask, cheng2021per} models are increasingly important in agriculture for tasks like phenotyping \cite{weyler2024phenobench, vit2019length, wang2022convolutional}, monitoring \cite{ariza2024object, badgujar2024agricultural}, pest detection \cite{ahmad2022deep, chen2021smartphone}, yield estimation \cite{koirala2019deep, darwin2021recognition}, etc. Due to the vast diversity of the agricultural domain, unique tasks and imaging setups, these models need dedicated, extensive and varied training data to ensure reliable performance. However, manual annotation of such datasets is time-consuming and costly, creating a major bottleneck in developing effective AI solutions. Streamlining or automating the annotation process would significantly improve the efficiency and scalability of these models, benefiting the agricultural sector.

Recent advances \cite{cheng2024yolo, du2022learning, gu2021open, liu2024grounding, shi2023edadet, wu2023aligning} have focused on using large vision-language models for open-set object detection in computer vision. These models are designed to detect objects beyond a predefined set by using human language input, making them highly versatile. Notable among these is Grounding-DINO\cite{liu2024grounding}, which has been trained on very large datasets including O365\cite{shao2019objects365}, OI\cite{openimages}, GoldG\cite{kamath2021mdetr}, Cap4M\cite{li2022grounded}, COCO\cite{lin2014microsoft}, and RefC\cite{kazemzadeh2014referitgame}. Building on this foundation, recent work \cite{ren2024grounded} incorporated SAM2\cite{ravi2024sam} to add segmentation capabilities, allowing the model to generate masks for detected objects.

A significant advantage of these models is their generalization ability, enabling them to detect unseen objects and adapt across various domains. Their large size, high computational cost, and slow inference speed pose challenges for real-time applications in agriculture. Despite these limitations, these models hold potential for automating annotation processes and distilling their knowledge into smaller, specialized models tailored for specific agricultural tasks. 

In this paper, we explore the application of pretrained Grounding-DINO on wide-range of publicly available agriculture datasets, including plant-weed detection \cite{haug2015crop, sb20}, insect identification \cite{sittinger2024insect}, wheat head detection \cite{david2023global}, fruit counting \cite{smitt2021pathobot, santos2020grape}, and remote sensing \cite{li2020object}. We further investigate instance segmentation using Grounding-DINO combined with SAM2 on datasets such as PhenoBench\cite{weyler2024phenobench}. 

First, we leverage zero-shot learning, where the model relies solely on text prompts corresponding to different classes without using any training images from the target datasets. However, we encounter several challenges: finding appropriate text prompts that consistently yield strong performance is difficult, particularly for objects like individual leaves in a plant due to overlapping structures. Additionally, distinguishing between visually similar classes, such as certain weeds resembling crop, proved challenging when generating text prompts. Finally, combining diverse-looking instances of the same class within a single text prompt adds further complexities.

To address the challenges associated with text prompting, we propose an efficient few-shot adaptation approach that enhances the performance of the Grounding-DINO model on agricultural datasets. Our method simplifies the original architecture by removing its language processing component (BERT) and introducing randomly initialized trainable parameters that mimic the shape of the BERT's text embeddings. This modification eliminates the need for text prompts, allowing us to adapt the model to specific datasets with minimal effort.

By learning these new text embeddings (a few thousand trainable parameters) with as few as two labeled images and iterating for a small number of training steps, we achieve excellent results across diverse agricultural datasets. This approach not only reduces the complexity of the model but also significantly speeds up the adaptation process for new datasets. We demonstrate superior performance compared to the zero-shot and other state-of-the-art few-shot approaches. This makes it particularly valuable in automating annotation tasks within agriculture and other fields where efficiency is critical.

Our main contributions can be summarized as follows:
\begin{itemize}
\item We investigate the application of Grounding-DINO model across diverse agricultural datasets for object detection task, leveraging its zero-shot capabilities to detect objects without requiring any labeled training data.
\item We discover manual prompt tuning for agriculture applications is impractical due to significant challenges in crafting effective prompts for zero-shot approach in various scenarios.
\item We propose a simple and efficient few-shot learning method for Grounding-DINO that eliminates the need for text prompts and enables efficient adaptation to new datasets using only a minimal number of training examples and iterations.
\item We conduct a series of experiments across various agricultural datasets, demonstrating our few-shot approach's superior performance compared to zero-shot and other state-of-the-art few-shot methods.
\end{itemize}

    \section{Related Work}
\label{sec:related_work}
\noindent\textbf{Object detection in Agriculture.} Object detection plays a pivotal role in agriculture for tasks \cite{ARIZASENTIS2024108757} including disease detection \cite{9044330}, crop identification \cite{Wang_2024_CVPR}, pest detection \cite{TANG2023102340}, fruit counting \cite{SANTOS2020105247}, and cattle tracking \cite{10855392}. Current approaches often rely on supervised learning frameworks, fine-tuning benchmark models like YOLO variants \cite{badgujar2024agricultural} for real-time performance.  However, these methods demand extensive annotated datasets, increasing costs and time requirements. They also exhibit limited generalization to novel classes due to their 
closed-set detection focus \cite{9093355}.


\noindent\textbf{Open-set Object Detection.} Open-set object detection extends traditional methods like Faster R-CNN \cite{NIPS2015_14bfa6bb} and YOLO \cite{7780460} to address novel object detection, building on Scheirer et al.'s \cite{6365193} open-set recognition framework. Recent advancements include Grounding-DINO \cite{liu2024grounding}, which leverages the DINO transformer and grounded pre-training for text-visual alignment, and YOLO-UniOW \cite{liu2024yolo}, demonstrating iterative vocabulary expansion to improve detection robustness. However, these models often struggle with agricultural datasets. To address this challenge, we opted for an efficient few-shot learning approach.


\noindent\textbf{Few-shot Object Detection.} 
Few-shot learning for object detection \cite{Han_2024_CVPR, XIN2024102307} aims to detect and classify objects in images with only a few annotated examples per class, addressing the challenge of misalignment with target region of interest \cite{pan2024solution} and data scarcity \cite{10.1145/3519022}. 
This is reinforced by the boost in detection performance on small agricultural datasets \cite{fsufs2022}. Our paper demonstrates the benefits of leveraging few-shot learning over a zero-shot foundational detection model like Grounding-DINO.

\noindent\textbf{Prompt Learning.} Text input (prompt) plays an important role in vision-language models, but finding the right prompt can be a challenging process. To address this challenge, recent works \cite{shin2020autoprompt, lester2021power} have proposed prompt learning techniques to systematically improve prompt design. Prompt tuning has been mainly explored with CLIP models \cite{radford2021learning} for image classification \cite{zhou2022conditional, zhou2022learning, kan2023knowledge}. For classification problems, given the template `a photo of \{label\},’ these methods utilize backpropagation to optimize the \{label\} token to match text and image features.
Recent work by Li et al. \cite{li2022grounded} employs prompt learning for object detection using the GLIP model, where text prompts (e.g., `detect fish') are utilized with the language encoder to generate text embeddings, which are then optimized using a few training images. Although, our approach is similar in concept, it differs in three key aspects: (1) we completely eliminate the use of text prompts for initialization and remove the text encoder (BERT) entirely; (2) our method initializes embeddings randomly and experiments with varying numbers of embeddings per class, demonstrating that performance improves as the number of embeddings per class increases; and (3) we apply this approach to pretrained Grounding-DINO model and benchmark it on agriculture-related datasets.
    \section{Method}
\label{sec:method}

\begin{figure*}[ht]
\vspace{-.25in}
  \centering
  \hspace{0.5in}
  \begin{minipage}[b]{0.34\textwidth}
  
    \centering
    \includegraphics[width=\linewidth]{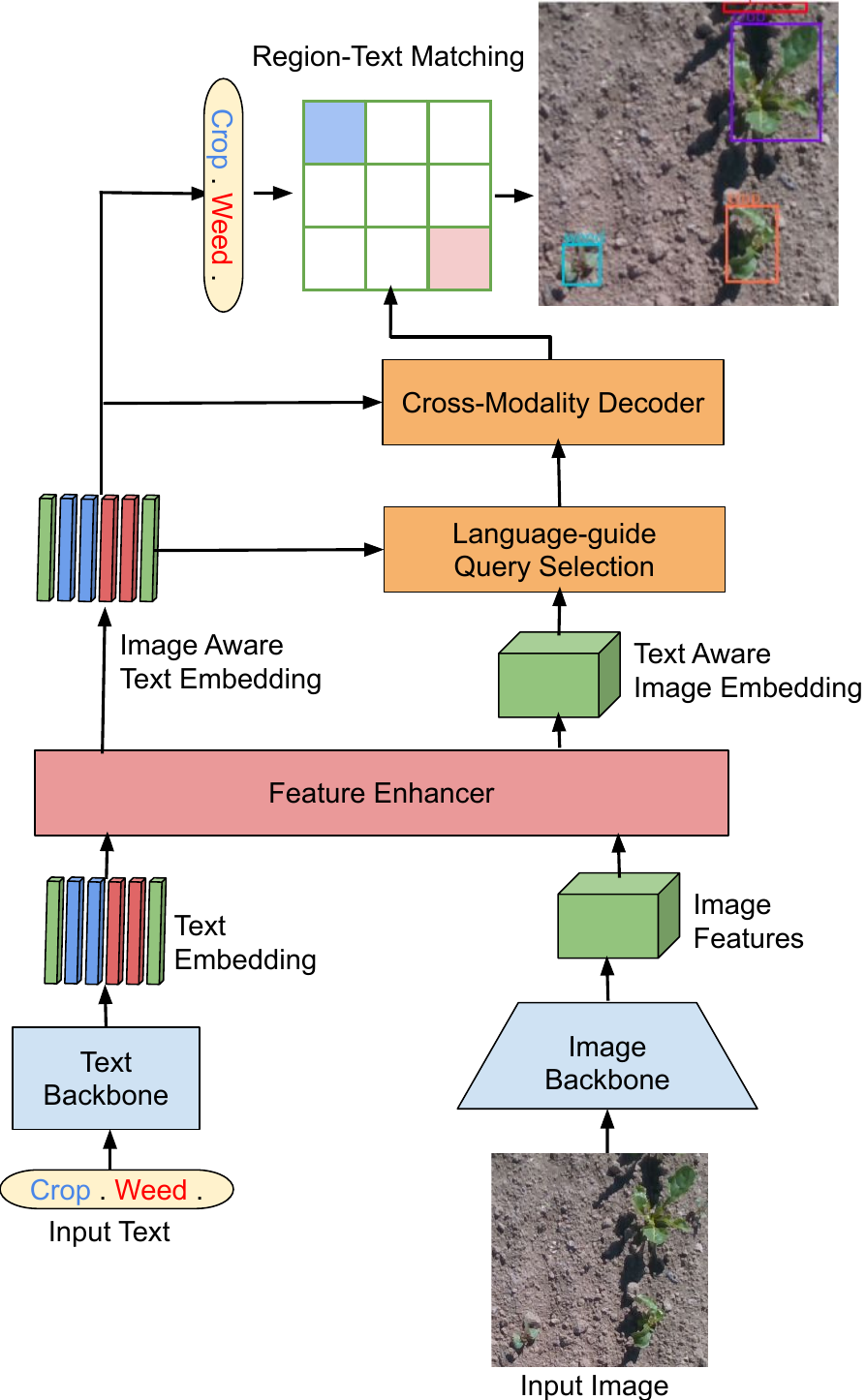}
    \vspace{-0.22in}
    \caption*{Grounding-DINO}
  \end{minipage}
  \hspace{-0.5in}
  \hfill
  \vrule width 1pt
  \hfill
  \hspace{-0.5in}
  \begin{minipage}[b]{0.34\textwidth}
    \centering
    \includegraphics[width=\linewidth]{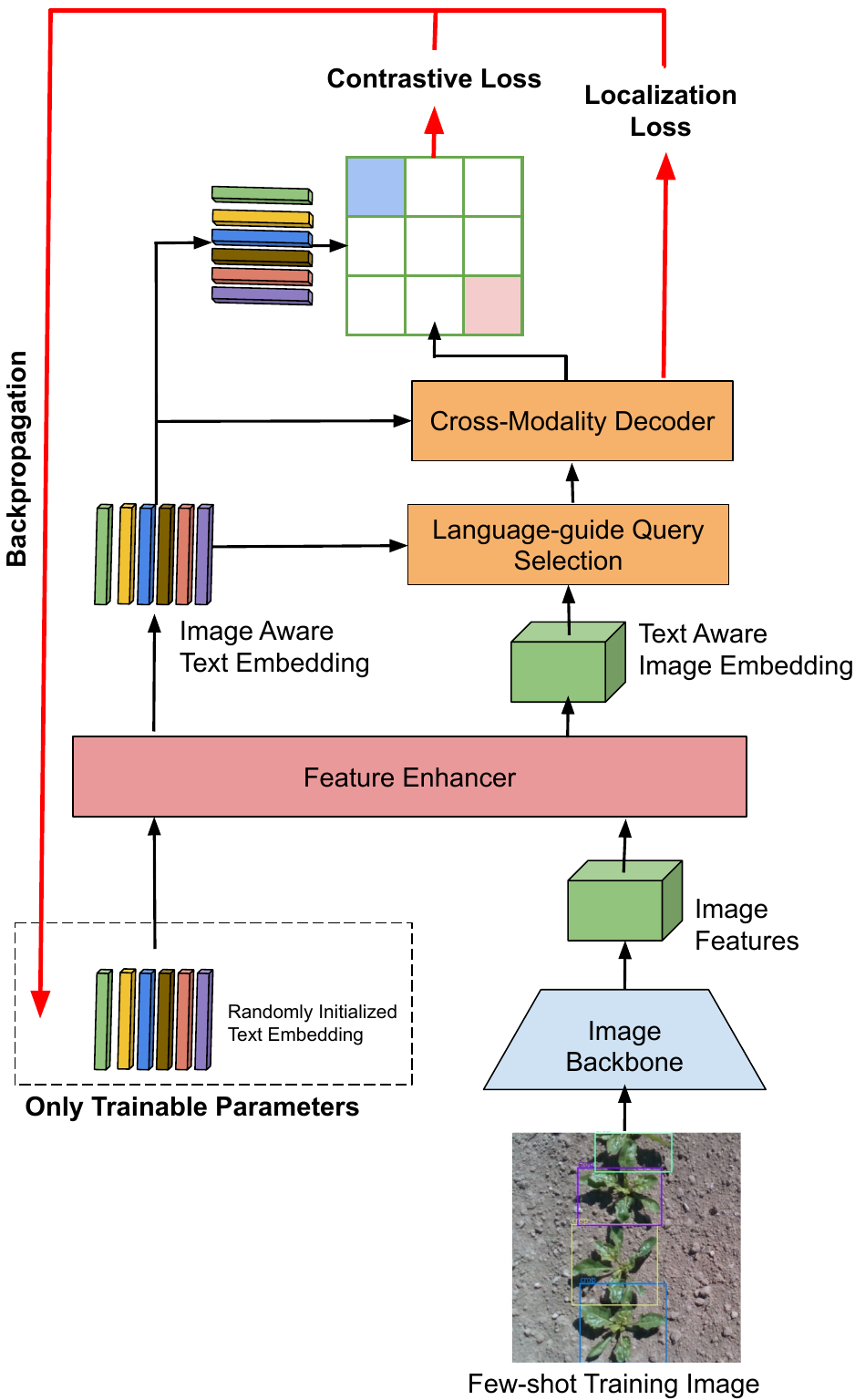}
    \vspace{-0.22in}
    \caption*{Proposed Few-shot Approach}
  \end{minipage}
  \vspace{-0.1in}
  \caption{Figure shows the block diagram of zero-shot inference with Grounding-DINO (left) and our proposed few-shot approach (right). Our method eliminates the BERT text encoder and operates directly in BERT's output space. Text embeddings are initialized randomly with dimensions matching BERT's outputs. We train only these text embeddings (approximately a few thousand parameters), keeping the rest of Grounding-DINO's parameters frozen, which requires a few labeled images and training iterations to achieve strong performance across diverse agricultural datasets.}
  \label{fig:block_diagram}
  \vspace{-0.2in}
\end{figure*}

In this section, we present our few-shot adaptation approach using Grounding-DINO for agricultural datasets. Our method leverages a pre-trained Grounding-DINO model and eliminates the need for text prompts, enabling direct optimization within its text feature space. This modification allows for efficient adaptation to various datasets with minimal effort. We provide detailed insights into our approach in the following subsections.

\subsection{Grounding-DINO}
Grounding-DINO is a popular large vision-language model equipped with open-set object detection capabilities. This enables the recognition of novel objects outside the initially defined categories.

As shown in Figure \ref{fig:block_diagram}, the Grounding-DINO model processes an image along with a text prompt corresponding to each class. The model comprises several key components: an image backbone that extracts image features, a text backbone that encodes textual information into feature vectors, a feature enhancer module that fuses these image and text features, a language-guided query selection module that initializes object queries based on the input text prompt, and finally, a cross-modality decoder that refines object features and bounding boxes.

For each (image, text) pair, the model processes images using the Swin Transformer\cite{liu2021swin} and text via a BERT encoder \cite{devlin2019bert}. Image features are hierarchically extracted from different layers of the Swin Transformer, capturing hierarchical visual information. Text is tokenized using the byte-pair encoding (BPE) scheme \cite{sennrich2015neural}, which is then encoded by the BERT model to produce $N\times768$-dimensional text embeddings (features), where $N$ represents the number of tokens in the text prompt.
 These raw image and text embeddings are fed into the feature enhancer module to enable cross-modal fusion. The module comprises multiple enhancer layers, each containing self-attention mechanisms for both image and text processing, as well as cross-attention layers that facilitate interactions between text-to-image and image-to-text contexts.

Grounding-DINO method uses the enhanced text features to select object queries by calculating the dot product between text and image features. It then selects $N_I$ object queries by choosing image features with the maximum scores. These language-guided queries are subsequently fed into a cross-modality decoder. In this decoder, query features undergo a series of self-attention operations, followed by cross-attentions with both image and text features. Finally, the output queries from the decoder's last layer are utilized to predict object boxes and corresponding class probabilities based on similarity with the text features.

For training, the model uses a contrastive loss between predicted objects and language token features. To compute this loss, each query calculates a dot product with text features to produce logits for every text token. Then the focal loss \cite{lin2017focal} is applied to these logits to obtain the classification loss. Additionally, L1 and GIoU\cite{rezatofighi2019generalized} losses are employed for the bounding box regression. Similar to DETR-like models \cite{carion2020end}, bipartite matching is used to find the matching between predicted and ground-truth objects. The total loss is then computed using the combination of classification and bounding-box losses based on this mapping.

\subsection{Zero-shot approach}
In the zero-shot setting with Grounding-DINO, the pre-trained model is evaluated on test sets from various datasets without fine-tuning. To construct text prompts for different classes, two approaches are employed: using single words separated by full stops (e.g., ``crop . weed .") and creating phrases for each class separated by full stops (e.g., ``green pepper . red pepper .") to enhance the distinction between similar categories. During class prediction, the model computes a dot product between text token features and predicted object features, identifying the highest score index. If this index corresponds to tokens within the class's text prompt set, the associated class is assigned.

\subsection{Few-shot approach}
In agricultural datasets, creating effective text prompts is particularly challenging due to their vastness and diversity, as well as the domain-specific nuances. To tackle this prompting challenge, we introduce a few-shot learning approach, where a small number of labeled images are used to adapt the pretrained grounding model for new datasets. This method is especially beneficial when labeled data is limited and enables a single pre-trained model to effectively handle diverse agricultural datasets by leveraging minimal examples for adaptation.

As shown in Figure \ref{fig:block_diagram}, our approach eliminates the BERT-based text encoder (dashed box in the right panel) and operates directly within the output space of the BERT model. Text embeddings are initialized randomly with dimensions matching BERT outputs, and position IDs along with attention masks are designed to ensure class-specific feature attention, mirroring BERT's mechanisms.

Similar to text prompts, a single class can have multiple words or tokens; in our setup, we use multiple embeddings per class. To maintain same shape as typical BERT model outputs, we include dummy embeddings for start and end tokens. Given C classes and T embeddings per class, the dimensionality of these text embeddings is $(C*T + 2)\times768$, where $+2$ accounts for the start-end tokens and $768$ represents the output dimension of the BERT model.

In our few-shot implementation, we fine-tune only the text embeddings while keeping all other parameters of the Grounding-DINO frozen. For a labeled image, the model outputs both class probabilities and bounding box coordinates.
Let $N_I$ denote the number of object queries with feature vector $X_I$, and let $N_T$ represent the number of text embeddings with feature vector $X_T$. The output class probability matrix is computed as

\[
P_{out} = \text{sigmoid}(X_I X_T^T)
\]

\noindent where $P_{out}$ has dimensions $N_I \times N_T$. For ground-truth probabilities $P_{gt}$, we assign a value of 1 for all token indices corresponding to the correct class and 0 otherwise.
We use focal loss \cite{lin2017focal} for classification, denoted by $L_{cls}$. For bounding box loss, we use $L1$ and GIoU loss \cite{rezatofighi2019generalized}, denoted by $L_{giou}$. Our total loss is given by

\[
L = \lambda_1 L_{cls} + \lambda_2 L1 + \lambda_3 L_{giou}
\]

\noindent where $\lambda_1$, $\lambda_2$, and $\lambda_3$ are hyper-parameters, empirically set to 1, 5, and 2, respectively. To match predicted objects $y$ with ground-truth objects $\hat{y}$, we use bipartite matching \cite{carion2020end}. The optimal assignment $\hat{\sigma}$ is determined by minimizing

\[
\hat{\sigma} = \operatorname*{arg\,min}_{\sigma} \sum_{i=1}^{N} L(y_i, \hat{y}_{\sigma(i)})
\]

Based on this optimal matching, we compute the final loss and update the trainable text embedding parameter $\mathcal{W}$ using backpropagation

\[
\mathcal{W}_{t+1} = \mathcal{W}_t - \eta \nabla_{\mathcal{W}} L_{\hat{\sigma}}(y, \hat{y})
\]

\noindent where $\eta$ is the learning rate and $\nabla_{\mathcal{W}}$ represents the gradient of the loss function with respect to $\mathcal{W}$. After optimizing the text embedding parameters using a few training images for a few training iterations, ${t}$, we obtain an optimally tuned embedding vectors. These learned embeddings are used for inference on test data.

For text feature initialization, we employ a normal distribution. While these features can be initialized using text prompts input through the BERT model, our experiments show that random initialization achieves comparable performance. Unlike text prompt initialization, which needs identifying meaningful phrases for each class, random initialization simplifies the process by eliminating this requirement. This approach not only simplifies the process but also removes the dependency on the BERT model from our pipeline, thereby saving some computational resources.


    \section{Experiments}
\label{sec:experiments}

\begin{figure*}[ht]
\vspace{-0.25in}
  \centering
  \begin{minipage}[b]{0.7\textwidth}
    \centering
    \includegraphics[width=\linewidth]{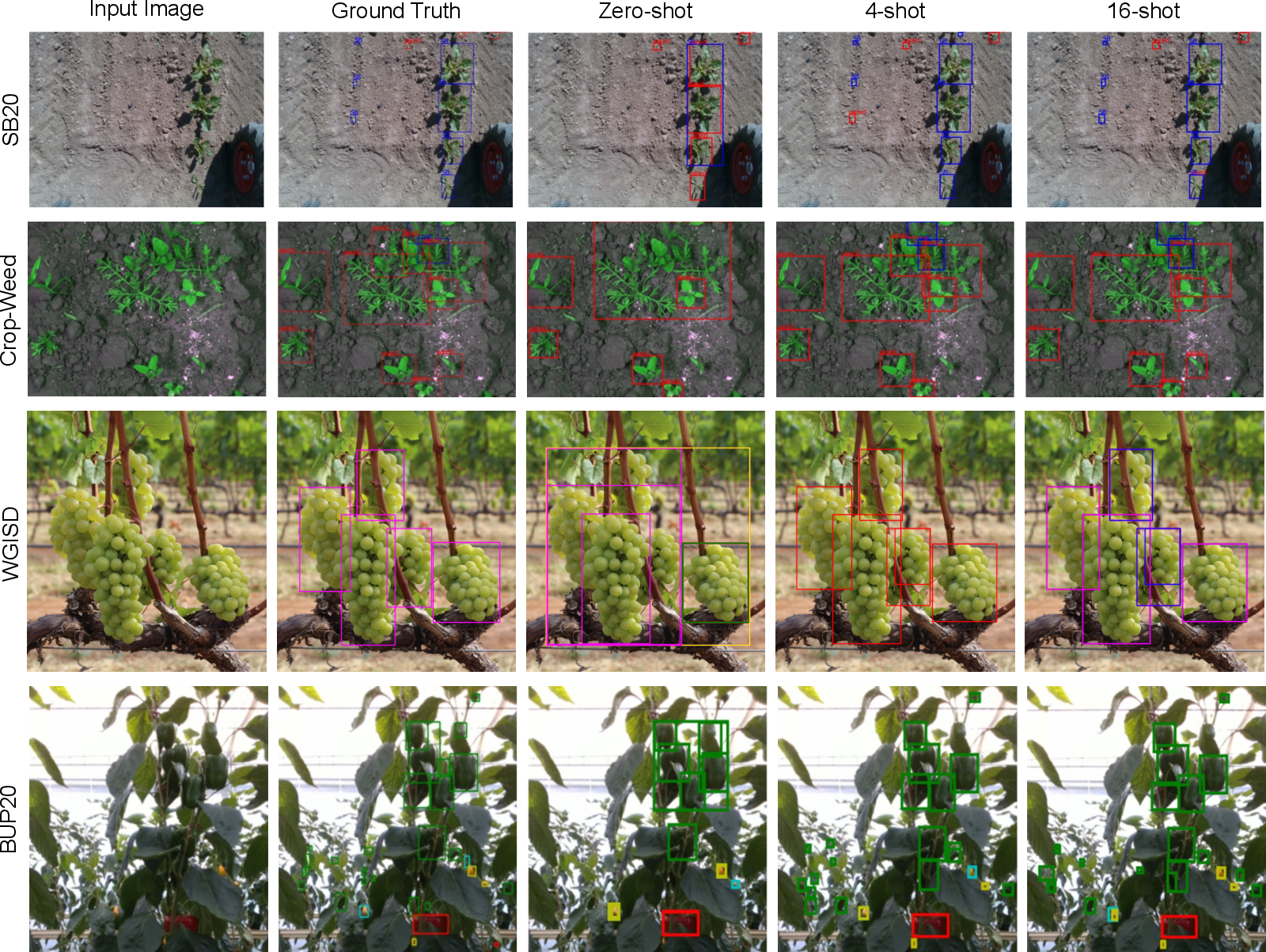}
  \end{minipage}
  \vspace{-0.11in}
  \caption{Figure compares zero-shot vs our few-shot approaches using Grounding-DINO on SB20\cite{sb20}, Crop-Weed\cite{haug2015crop}, Grape Detection (WGISD)\cite{santos2020grape}, and BUP20\cite{smitt2021pathobot}. Zero-shot fails in cluttered/occluded environments, struggling to detect instances or distinguish similar classes. Our few-shot approach (4-shot and 16-shot) outperforms zero-shot on all datasets.}
  \label{fig:datasets}
  \vspace{-0.11in}
\end{figure*}
In this section, we present our implementation details, evaluation metrics, and datasets. We use a pretrained Grounding-DINO base model \cite{liu2024grounding} for all experiments. For zero-shot detection, we employ text prompts that best describe the dataset classes (either single words or phrases). In our few-shot experiments on target datasets, we use multiple embeddings per class, with an ablation study showing that 4 embeddings per class yield good performance. 

For all experiments, we initialize text embeddings randomly from a normal distribution and train these embeddings for $400$ iterations with a batch size of $4$ on an Nvidia A100 GPU. We use an initial learning rate of $2.0$ with cosine decay and AdamW optimizer. Each image is resized by scaling its shorter side to $800$ pixels while maintaining the aspect ratio, and we apply random horizontal flips and random cropping as training augmentations.
For few-shot training, unless otherwise noted, we randomly select $1$, $2$, $4$, $8$, $16$, and $24$ images from the training set of each target dataset and evaluate on complete test set. For consistency in our experiments, `shots' are referred to as the number of training images used. We perform $10$ runs for each few-shot setup by sampling random few-shot training images and report mean and standard deviation scores.

We evaluate model performance using object detection metrics: Mean Average Precision (mAP) at Intersection over Union (IoU) thresholds of $50$:$95\%$, $50\%$, and $75\%$. Unless otherwise stated, mAP refers to mAP@$50$:$95$ in our paper. We benchmark the model on $8$ datasets, as detailed in the following subsection.

\subsection{Object Detection}

\noindent\textbf{Crop-Weed Dataset \cite{haug2015crop}:}
This dataset is designed to evaluate computer vision models for precision agriculture tasks. It contains top-down field images captured by an autonomous field robot in an organic carrot farm during various crop growth stages, with images taken when one or more true leaves are present. The dataset includes two classes: crop (carrot) and weed. This dataset presents unique challenges due to high similarity between weed plants and young carrot plants, as well as frequent partial occlusion of carrots by weeds, as shown in Figure \ref{fig:datasets}.

First, we perform an ablation study to determine the optimal number of text embeddings per class. We use the same $4$ and $8$ training images and increase the number of text embeddings while keeping other parameters same. Table \ref{table:tokens_comparison} shows that performance improves as we increase the number of text embeddings per class; however, it plateaus after $4$ embeddings per class. We observe similar trends in both 4-shot and 8-shot settings. For consistency across experiments, we choose $4$ embeddings per class for all datasets.

Next, we compare the performance of our few-shot approach against the zero-shot baseline. For the zero-shot setup, we tested various prompts for the carrot plant class, such as `small carrot plant' and `carrot plant.' However, the model frequently misclassifies instances, often detecting all instances simply as `plant' or `weed,' as shown in Figure \ref{fig:datasets}. We report the zero-shot results for the prompt `crop . weed' in Table \ref{table:crop_weed}. From these results, we observe that the few-shot approach significantly outperforms the zero-shot method. With $24$ images, our few-shot approach achieves an mAP of $43.0$, outperforming zero-shot detection at $10.5$.


\begin{table}[ht!]
\vspace{-0.1in}
    \centering
    \captionsetup{width=.99\linewidth}
    \caption{Comparison of mAP for different number of text embeddings per class on Crop-Weed dataset \cite{haug2015crop}. Here number of shot is number of training images used.}
\vspace{-0.1in}
    
    \label{table:tokens_comparison}
    \begin{small}
    \resizebox{0.48\textwidth}{!}{\begin{tabular}{c|cc|ccc}
        \toprule
        \textbf{Number of} & \multicolumn{2}{c|}{\textbf{4-Shot}} & \multicolumn{2}{c}{\textbf{8-Shot}} \\
        \textbf{Embeddings} & \textbf{mAP@(50:95)} & \textbf{mAP@50}  & \textbf{mAP@(50:95)} & \textbf{mAP@50} \\
        \midrule
        2 & 39.5 & 62.8  & 39.2 & 64.0  \\
        4 & 42.6 & 67.7  & 42.6 & 68.3  \\
        6 & 41.5 & 67.2  & 42.1 & 68.1  \\
        8 & 42.1 & 68.7  & 42.2 & 68.4  \\
        10 & 42.2 & 68.7  & 42.7 & 68.9  \\
        \bottomrule
    \end{tabular}}
    \end{small}
\vspace{-0.1in}
\end{table}

\begin{table*}[ht!]
\caption{Quantitative results on various datasets comparing our few-shot approach with zero-shot. Our few-shot method significantly outperforms zero-shot across all datasets, with performance improving as we increase the number of training images.}
\label{table:6_map_box}
\vspace{-0.07in}
\parbox{.32\textwidth}{
    \centering
    \subcaption{Crop-Weed}
    \label{table:crop_weed}
    \begin{small}
    \resizebox{0.32\textwidth}{!}{\begin{tabular}{cccc}
        \toprule
         \textbf{Number of} & \textbf{mAP} & \textbf{mAP} & \textbf{mAP} \\
        \textbf{training images} &\textbf{@(50:95)} & \textbf{@50} & \textbf{@75} \\
        \midrule
        0 & 10.5 & 26.8 & 6.6 \\
        \hline
        1 & 22.7 $\pm$ 7.0 & 39.1 $\pm$ 12.4 & 24.0 $\pm$ 6.8 \\
        2 & 31.5 $\pm$ 4.1 & 53.0 $\pm$ 6.2  &  31.3$\pm$5.4 \\
        4 & 36.9 $\pm$ 3.3 & 61.9 $\pm$ 5.2  &  38.7$\pm$4.4\\
        8 & 39.5 $\pm$ 1.6 & 64.6 $\pm$ 2.6  &   42.8$\pm$3.2\\
        16 & 42.5 $\pm$ 2.1 & 68.0 $\pm$ 3.3  &  46.6 $\pm$ 2.8\\
        24 & \textbf{43.0 $\pm$ 3.5} & \textbf{68.6 $\pm$ 4.9}  &  \textbf{48.7 $\pm$ 4.8}\\
        \bottomrule
    \end{tabular}
    }
    \end{small}
}
\hfill
\parbox{.32\textwidth}{
    \centering
    \subcaption{SB20}
    \label{table:SB20}
    \begin{small}
    \resizebox{0.32\textwidth}{!}{\begin{tabular}{cccc}
        \toprule
         \textbf{Number of} & \textbf{mAP} & \textbf{mAP} & \textbf{mAP} \\
        \textbf{training images} &\textbf{@(50:95)} & \textbf{@50} & \textbf{@75} \\
        \midrule
        0  & 26.5 & 38.6 & 28.9 \\
        \hline
        1  & 19.6 $\pm$ 7.1 & 33.8 $\pm$ 12.2 & 19.6 $\pm$ 7.9 \\
        2  & 29.4 $\pm$ 7.7 & 50.3 $\pm$ 12.3 & 29.6 $\pm$ 8.4 \\
        4  & 37.7 $\pm$ 6.2 & 61.7 $\pm$ 8.2 & 38.3 $\pm$ 7.0 \\
        8  & 41.9 $\pm$ 5.8 & 67.8 $\pm$ 7.7 & 42.7 $\pm$ 6.5 \\
        16 & 45.6 $\pm$ 1.8 & 72.8 $\pm$ 2.3 & 46.9 $\pm$ 2.0 \\
        24 & \textbf{46.4 $\pm$ 2.0} & \textbf{73.3 $\pm$ 2.6} & \textbf{47.6 $\pm$ 2.2 }\\
        \bottomrule
    \end{tabular}
    }
    \end{small}
}
\hfill
\parbox{.32\textwidth}{
    \centering
    \subcaption{PhenoBench}
    \label{table:phenobench}
    \begin{small}
    \resizebox{0.32\textwidth}{!}{
    \begin{tabular}{cccc}
        \toprule
         \textbf{Number of} & \textbf{mAP} & \textbf{mAP} & \textbf{mAP} \\
        \textbf{training images} &\textbf{@(50:95)} & \textbf{@50} & \textbf{@75} \\
        \midrule
        0  & 12.6 & 23.2 & 12.4 \\
        \hline
        1  & 27.8 $\pm$ 6.7 & 53.4 $\pm$ 9.3 & 25.4 $\pm$ 8.0 \\
        2  & 32.8 $\pm$ 1.4 & 61.4 $\pm$ 1.6 & 30.8 $\pm$ 2.2 \\
        4  & 35.3 $\pm$ 1.3 & 65.4 $\pm$ 1.5 & 33.8 $\pm$ 1.9 \\
        8  & 37.4 $\pm$ 0.5 & 67.9 $\pm$ 1.0 & 36.3 $\pm$ 0.7 \\
        16 & 37.3 $\pm$ 1.7 & 67.7 $\pm$ 2.2 & 36.0 $\pm$ 2.0 \\
        24 & \textbf{38.2 $\pm$ 1.0} & \textbf{68.7 $\pm$ 1.3} & \textbf{37.1 $\pm$ 1.4} \\
        \bottomrule
    \end{tabular}
    }
    \end{small}
}
\vfill
\parbox{.32\textwidth}{
    \centering
    \subcaption{Insect Detection}
    \label{table:insect}
    \begin{small}
    \resizebox{0.32\textwidth}{!}{
    \begin{tabular}{cccc}
        \toprule
         \textbf{Number of} & \textbf{mAP} & \textbf{mAP} & \textbf{mAP} \\
        \textbf{training images} &\textbf{@(50:95)} & \textbf{@50} & \textbf{@75} \\
        \midrule
        0  & 16.0 & 25.0 & 19.1 \\
        \hline
        1  & 6.5 $\pm$ 1.6 & 10.2 $\pm$ 2.3 & 6.8 $\pm$ 1.7 \\
        2  & 15.1 $\pm$ 2.1 & 23.2 $\pm$ 4.0 & 16.6 $\pm$ 2.4 \\
        4  & 20.8 $\pm$ 5.8 & 30.5 $\pm$ 8.2 & 23.5 $\pm$ 6.8 \\
        8  & 30.8 $\pm$ 4.8 & 45.0 $\pm$ 7.7 & 34.7 $\pm$ 5.1 \\
        16 & 39.0 $\pm$ 4.6 & 56.1 $\pm$ 6.3 & 44.0 $\pm$ 5.2 \\
        24 & \textbf{41.0 $\pm$ 5.0} & \textbf{59.0 $\pm$ 6.6} & \textbf{46.3 $\pm$ 5.7} \\
        \bottomrule
    \end{tabular} 
    }
    \end{small}
}
\hfill   
\parbox{.32\textwidth}{
    \centering
    \subcaption{Grape Detection}
    \label{table:grape}
    \begin{small}
    \resizebox{0.32\textwidth}{!}{
    \begin{tabular}{cccc}
        \toprule
        \textbf{Number of} & \textbf{mAP} & \textbf{mAP} & \textbf{mAP} \\
        \textbf{training images} &\textbf{@(50:95)} & \textbf{@50} & \textbf{@75} \\
        \midrule
        0  & 3.8 & 6.1 & 4.3 \\
        \hline
        1  & 2.9 $\pm$ 1.0 & 4.3 $\pm$ 1.2 & 3.0 $\pm$ 1.1 \\
        2  & 9.7 $\pm$ 2.9 & 14.5 $\pm$ 4.4 & 10.5 $\pm$ 3.3 \\
        4  & 22.6 $\pm$ 7.6 & 32.9 $\pm$ 11.2 & 24.7 $\pm$ 8.6 \\
        8  & 32.1 $\pm$ 8.0 & 46.5 $\pm$ 11.7 & 34.7 $\pm$ 8.7 \\
        16 & \textbf{35.8 $\pm$ 9.2} & \textbf{51.2 $\pm$ 13.5} & \textbf{39.0 $\pm$ 10.1} \\
        24 & 35.4 $\pm$ 7.0 & 49.9 $\pm$ 9.9 & 39.0 $\pm$ 7.9 \\
        \bottomrule
    \end{tabular} 
    }
    \end{small}
}  
\hfill
\parbox{.32\textwidth}{
    \centering
    \subcaption{Wheat Head Detection}
    \label{table:wheat_head}
    \begin{small}
    \resizebox{0.32\textwidth}{!}{
        \begin{tabular}{cccc}
        \toprule
         \textbf{Number of} & \textbf{mAP} & \textbf{mAP} & \textbf{mAP} \\
        \textbf{training images} &\textbf{@(50:95)} & \textbf{@50} & \textbf{@75} \\
        \midrule
        0  & 10.2 & 22.8 & 7.5 \\
        \hline
        1  & 31.4 $\pm$ 4.1 & 74.0 $\pm$ 6.5 & 20.5 $\pm$ 4.6 \\
        2  & 34.8 $\pm$ 3.3 & 79.9 $\pm$ 4.2 & 23.5 $\pm$ 3.8 \\
        4  & 37.1 $\pm$ 2.6 & 83.3 $\pm$ 3.4 & 26.1 $\pm$ 3.0 \\
        8  & 37.9 $\pm$ 1.6 & 84.7 $\pm$ 1.8 & 26.6 $\pm$ 2.0 \\
        16 & 39.6 $\pm$ 0.8 & 86.4 $\pm$ 0.8 & 29.0 $\pm$ 1.1 \\
        24 & \textbf{40.1 $\pm$ 1.0} & \textbf{86.8 $\pm$ 0.9} & \textbf{29.9 $\pm$ 1.5} \\
        \bottomrule
    \end{tabular}
    }
    \end{small}
}      
\vspace{-0.2in}
\end{table*}

\noindent\textbf{BUP20 Dataset \cite{smitt2021pathobot}:}
This dataset includes images of sweet peppers across five classes: red, yellow, green, mixed red, and mixed yellow. It is challenging dataset due to significant occlusion by green leaves as shown in Figure \ref{fig:datasets}.

\begin{table}[ht!]
\vspace{0.02in}
    \centering
    \captionsetup{width=.99\linewidth}
    \caption{Comparison of our few-shot approach against YOLOv11's performance on the BUP20 dataset \cite{smitt2021pathobot}, where the entire YOLOv11 model is trained using few training images.}
    \label{table:bup20}
    \vspace{-0.1in}
    \begin{small}
    \resizebox{0.48\textwidth}{!}{\begin{tabular}{c|cc|cc}
        \toprule
        \textbf{Number} & \multicolumn{2}{c|}{\textbf{Few-shot}} & \multicolumn{2}{c}{\textbf{YOLOv11}} \\
        \textbf{of Images} & \textbf{mAP@(50:95)} & \textbf{mAP@50}  & \textbf{mAP@(50:95)} & \textbf{mAP@50}  \\
        \midrule
        0  & 21.7 & 32.2 & - & - \\
        \hline
        1  & 17.5 $\pm$ 3.3 & 26.8 $\pm$ 4.4 & 2.9 $\pm$ 0.5 & 4.7 $\pm$ 0.7 \\
        2  & 23.1 $\pm$ 2.4 & 34.8 $\pm$ 3.5 & 3.5$\pm$ 0.8 & 5.7$\pm$ 1.2 \\
        4  & 28.1 $\pm$ 2.7 & 42.1 $\pm$ 3.5 & 4.6$\pm$0.5 & 7.6$\pm$0.9 \\
        8  & 32.1 $\pm$ 2.3 & 47.9 $\pm$ 3.5 & 12.5$\pm$0.8 & 19.8$\pm$1.2 \\
        16 & 36.5 $\pm$ 2.2 & 53.2 $\pm$ 3.0 & 18.4$\pm$1.7 & 29.1$\pm$2.5 \\
        24 & 38.1 $\pm$ 1.5 & 56.1 $\pm$ 2.1 & 21.8$\pm$2.0 & 34.3$\pm$2.7 \\        
        \bottomrule
    \end{tabular}}
    \end{small}
    \vspace{-0.15in}
\end{table} 



For zero-shot, we used the prompt `red pepper. yellow pepper. green pepper. mixed red pepper. mixed yellow pepper.' Table \ref{table:bup20} compares zero-shot, and few-shot performance with varying training images. With one image, few-shot learning underperforms compared to zero-shot, likely because not all class instances are represented in a single image. However, increasing training images improves performance significantly. With $24$ images, our few-shot approach achieves an mAP of $38.1,$ outperforming zero-shot detection at $21.7$.

We also compare the performance of our few-shot approach with YOLOv11\cite{Jocher_Ultralytics_YOLO_2023}. We use coco-pretrained YOLOv11-nano model and fine-tune the entire model parameters on the few training images for 100 epochs. Table \ref{table:bup20} demonstrate that the few-shot approach significantly surpasses YOLOv11, achieving an mAP improvement of up to $\sim 24\%$ when trained on just $4$ images.

\noindent\textbf{SB20 Dataset \cite{sb20}:}
This dataset contains RGB images of two classes: sugar beet and weeds. The images cover a range of growth stages, natural world illumination conditions, and challenging occlusions. 
For zero-shot detection, we tested prompts like `sugar beet' and `plant,' but the model struggled to distinguish between sugar beet and weeds as shown in Figure \ref{fig:datasets}. 

In Table \ref{table:SB20}, we use the prompt `crop . weed' for the zero-shot experiments. The table shows that with one training image, few-shot learning underperforms compared to zero-shot. This is likely due to the diverse growth stages of sugar beet plants and randomly sampling one training image fails to capture the full variation across all growth stages.
However, increasing training images improves performance significantly. With $24$ training images, our few-shot approach achieves an mAP of $46.4$, outperforming zero-shot detection at $26.5$. Additionally, the standard deviation decreases as we increase the number of training images. 

\noindent\textbf{Pheno Bench Dataset \cite{weyler2024phenobench}:}
This dataset comprises large high-resolution images captured with unmanned aerial vehicles (UAV) of sugar beet fields under natural lighting conditions over multiple days.
For our experiment, we focus on detecting individual leaves. This task is challenging due to the cluttered nature of the leaves and varying growth stages.

For zero-shot detection, we use the `leaf instance' prompt. However, as shown in Figure \ref{fig:teaser}, the model tends to detect entire rows or plants rather than individual leaves. Table \ref{table:phenobench} compares zero-shot and few-shot performance. We find that even with one training image, the few-shot approach outperforms zero-shot. With $24$ images, our few-shot method achieves an mAP of $38.2,$ significantly surpassing zero-shot detection at $12.6$.

\noindent\textbf{Insect Detection Dataset \cite{sittinger2024insect}:}
This dataset includes high-resolution images of various insects across six classes: fly, honeybee, hover fly, shadow, wasp, and other insect. It is challenging due to the small size of insects and similar-looking species, making differentiation difficult.

For zero-shot text prompting, this dataset presents challenges as the `other insect' class encompasses numerous species, complicating the creation of descriptive prompts that capture all variations within the class while distinguishing it from others. For quantification, we use the class names directly for prompting. Table \ref{table:insect} compares zero-shot and few-shot performance. We observe that few-shot detection with $1$ or $2$ images performs less effectively than zero-shot, as most dataset images contain insects from one or two classes, 
limiting model exposure to other species. However, increasing training images significantly improves performance; with $24$ images, our few-shot approach achieves an mAP of $41.0,$ outperforming zero-shot detection at $16.0.$

\noindent\textbf{Grape Detection Dataset \cite{santos2020grape}:}
It provides instances of five different grape varieties (Chardonnay, Cabernet Franc, Cabernet Sauvignon, Sauvignon Blanc, Syrah) captured under field conditions. These instances exhibit variance in grape pose, illumination as well as genetic and phenological characteristics such as shape, color, and compactness. 

For zero-shot prompting, we prepend `grape' to every class name. Figure \ref{fig:datasets} provides a qualitative comparison between zero-shot and few-shot detection. We observe that the zero-shot model struggles to detect individual clusters, often grouping them into one instead. Table \ref{table:grape} shows that for $1$ and $2$ training images, few-shot performance is suboptimal as these images typically contain only one or two classes each. However, at higher training numbers, our few-shot approach significantly outperforms zero-shot; for instance, with $16$ training images, it achieves $35.8$ mAP compared to $3.8$ for zero-shot.

\noindent\textbf{Wheat Head Dataset \cite{david2023global}:}
This dataset consists of images of wheat fields and contains a single class: `wheat spike head'. It is designed for counting wheat spike heads. The task presents significant challenges due to the presence of clutter and occluded heads.  

For zero-shot detection, we used the prompt `wheat spike head'. As shown in Figure \ref{fig:teaser}, the zero-shot model detects only a few heads and fails to detect occluded ones. Table \ref{table:wheat_head} shows that the few-shot approach significantly outperforms zero-shot detection, even with just one training image. With 24 training images, the few-shot method achieves $40.1$ mAP compared to $10.2$ for zero-shot.

\noindent\textbf{Remote Sensing Dataset \cite{li2020object}:}
Object detection in optical remote sensing (DIOR) dataset is a widely used benchmark for evaluating few-shot models in remote sensing. Collected from Google Earth, it spans over 80 countries, offering extensive variations in environmental conditions such as weather, seasons, and imaging quality. Our experiments are conducted across all splits of the dataset, each containing five classes. Following the experimental setup outlined in \cite{lin2025generalization}, we use the same images for each n-shot experiment. In this set up, the number of shots refers to the number of instances per class.

\begin{figure}
  \vspace{-0.25in}
    \centering
    \includegraphics[width=0.8\linewidth]{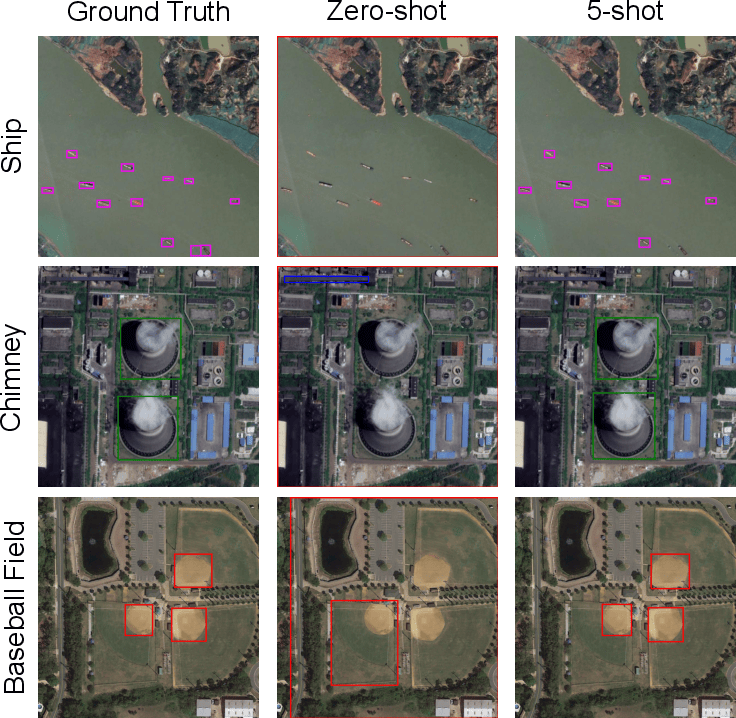}
      \vspace{-0.10in}
    \caption{Qualitative comparison between zero-shot and few-shot approaches on DIOR \cite{li2020object} dataset, demonstrating that our few-shot method achieves significantly better results.}
    \label{fig:dior}
      \vspace{-0.22in}
\end{figure}

Figure \ref{fig:dior} presents the qualitative results on DIOR Split-1, which contains 5 classes: baseball field, basketball court, bridge, chimney, and ship. For zero-shot detection, we use class names as prompts; Grounding-DINO achieves an mAP of $22.8$. We compare our few-shot approach with state-of-the-art (SOTA) methods in Table \ref{table:dior}. Our approach significantly outperforms previous methods: for 3-shot detection, we achieve an mAP of $40.0$, compared to the previous SOTA method's $31.69$. Consistent improvements are observed across other splits: on Split-2 for 3-shot, our approach achieves an mAP of $38.5$, compared to the previous SOTA's $14.5$; and on Split-3 for 3-shot, we attain an mAP of $30.0$, outperforming the previous SOTA's $18.85$.

\begin{table}[ht!]
\vspace{-0.1in}
    \centering
    \caption{Results of our few-shot approach and state-of-the-art few-shot model performance on DIOR dataset \cite{li2020object} (split-1).}
    \label{table:dior}
    \vspace{-0.1in}
    \begin{small}

    \begin{tabular}{cccccc}
        \toprule
        Method & 3-shot & 5-shot & 10-shot & 20-shot \\
        \midrule
        FSCE \cite{sun2021fsce} & 27.91 & 28.60 & 33.05 & 37.55 \\
        SAE-FSDet \cite{liu2024few} & 28.80 & 32.40 & 37.09 & 42.46 \\
        G-FSDet \cite{zhang2023generalized} & 27.60 & 29.89 & 34.86 & 37.49 \\
        GE-FSOD \cite{lin2025generalization} & 31.69 & 34.88 & 38.02 & 43.08 \\
        \hline
        \textbf{Ours} & \textbf{40.0} & \textbf{42.3} & \textbf{48.3} & \textbf{47.6}\\
        \bottomrule
    \end{tabular}
 \end{small}
 \vspace{-0.1in}
\end{table}

\subsection{Instance Segmentation}
\begin{figure}
  \vspace{-0.0in}
    \centering
    \includegraphics[width=0.8\linewidth]{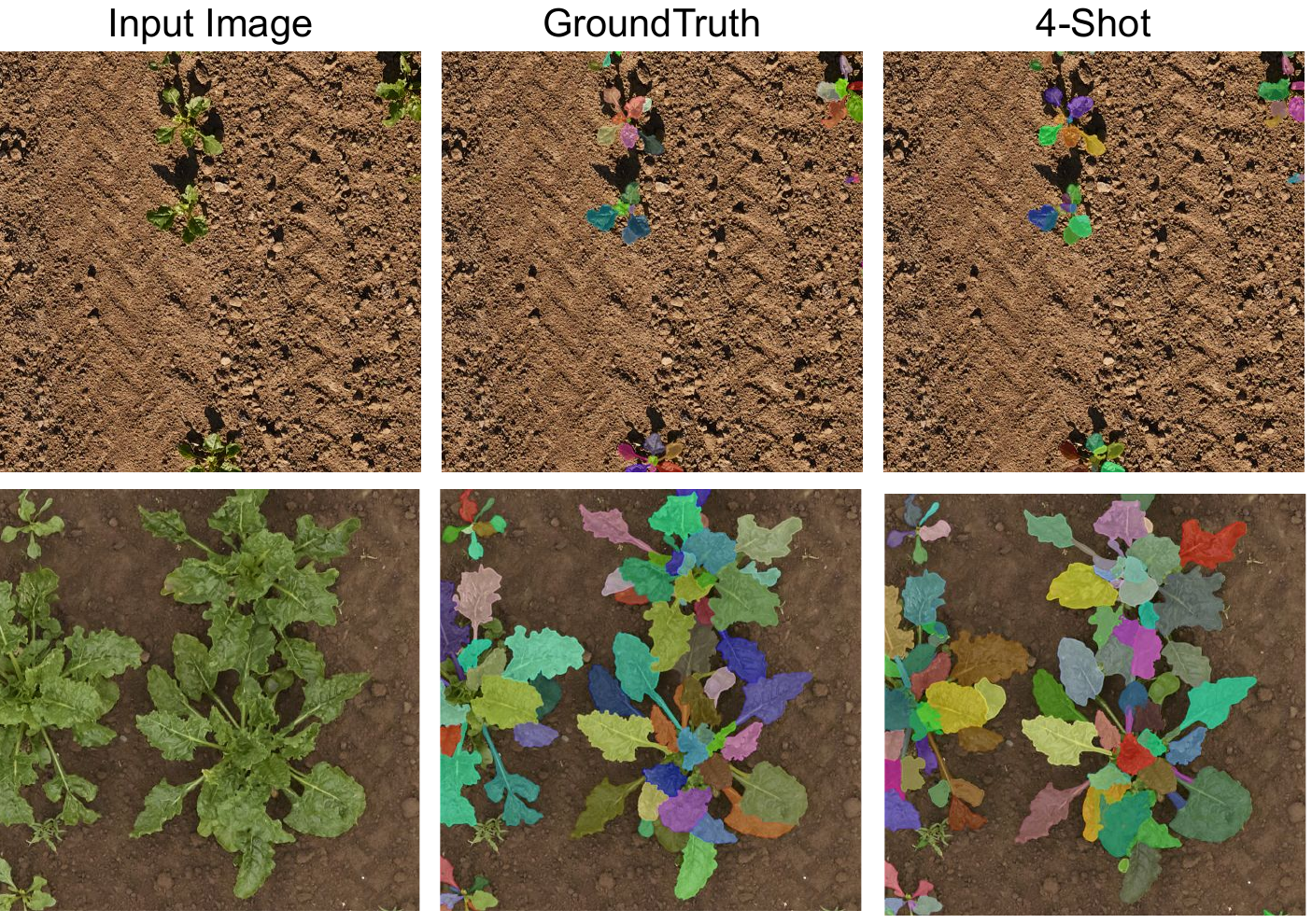}
      \vspace{-0.10in}
    \caption{Instance segmentation on PhenoBench dataset \cite{weyler2024phenobench} using our few-shot Grounding-DINO and SAM2.}
    \label{fig:pheno_bench_seg}
      \vspace{-0.2in}
\end{figure}

We extend our few-shot object detection approach to instance segmentation. Following \cite{ren2024grounded}, we use the SAM2 model \cite{ravi2024sam} as our foundation, using the few-shot predicted bounding boxes from our Grounding-DINO based detection method as prompts for SAM2. For experimentation, we utilize the PhenoBench dataset \cite{weyler2024phenobench}, where we perform leaf instance segmentation. Figure \ref{fig:pheno_bench_seg} demonstrates qualitative results on this dataset. The few-shot approach performs well in instance segmentation, particularly for plants at early growth stages, where leaves are less cluttered and the model achieves good performance. However, challenges arise with larger plants, as overlapping leaves lead to some segmentation errors due to occlusion. In terms of mAP with mask IoU, our model achieves $31.8$ with just one training image and improves to $42.3$ with eight training images.

\subsection{Discussion}

Our few-shot approach significantly outperforms the zero-shot method using Grounding-DINO model across diverse agricultural datasets, even with very limited training images and steps. Through our experiments (Table \ref{table:6_map_box}), we observe that the zero-shot Grounding-DINO model excels at detecting fruits and plants when they are visually sparse but struggles in cluttered or occluded scenarios, often failing to detect individual instances or grouping multiple instances together. Additionally, for multi-class fine-grained distinctions, such as crop-weed differentiation, the zero-shot approach proves inadequate in distinguishing between similar looking classes. These challenges stem from difficulties in formulating effective text prompts for individual instance detection and intra-class differentiation. Furthermore, datasets like insects, where multiple classes are grouped into one category, pose additional hurdles in crafting appropriate text prompts.

Our few-shot approach addresses the challenges of prompting and demonstrates significant improvements over the zero-shot method with minimal training images. Our experiments reveal that using just one training image for multi-class datasets, generally underperforms compared to zero-shot, as a single image often does not contain all class instances. However, as we increase the number of training images, the performance of our few-shot approach improves significantly. When comparing our few-shot approach to YOLO (Table \ref{table:bup20}), where the entire model parameters are fine-tuned, we observe that our method outperforms YOLO when the number of training images is limited. However, with all training images, YOLO's fully fine-tuned model performs better. Our future work would explore adding more trainable parameters to Grounding-DINO, such as through low-rank adapters, for scenarios with a larger number of training data. Additionally, when benchmarked against SOTA few-shot methods in remote sensing using standard experimental setups, our approach demonstrates significant improvements. This likely stems from the strong pre-training of Grounding-DINO on extensive datasets.
Furthermore, the pre-trained Grounding-DINO demonstrates robust adaptability across diverse datasets-from lab environments to satellite imagery-using minimal few-shot training examples. This versatility underscores its capacity to generalize seamlessly across modalities without domain-specific retraining, positioning it as a powerful tool for automating annotation pipelines in precision agriculture. While Grounding-DINO incurs notable computational costs, our proposed method can be seamlessly integrated into YOLO-based frameworks like YOLO-World\cite{cheng2024yolo} and YOLOE\cite{wang2025yoloe} for efficient few-shot detection tasks.

    \section{Conclusion}
\label{sec:conclusion}

In this paper, we investigate the application of the Grounding-DINO model for agricultural object detection tasks. We identified significant challenges in manual text prompting for agriculture-specific scenarios and developed an efficient few-shot learning method that eliminates reliance on text prompts. Our experimental results demonstrate superior performance across diverse datasets, particularly in challenging environments with occluded objects and fine-grained distinctions. Looking ahead, our work opens new avenues for research in methods for efficiently adapting open-set object detection models pretrained on large datasets and developing few-shot learning approaches tailored to agricultural datasets. The insights gained from this study contribute to the broader goal of advancing AI applications in agriculture, ultimately supporting more sustainable and efficient farming practices.
    \small
        \bibliographystyle{ieeenat_fullname}
        \bibliography{main}


\end{document}